\documentclass[letterpaper, 10 pt, conference]{ieeeconf}  

\IEEEoverridecommandlockouts                              

\usepackage{epsfig} 
\usepackage{mathptmx} 
\usepackage{times} 
\usepackage{amsmath} 
\usepackage{amssymb}  
\usepackage{multicol}
\usepackage{siunitx}
\usepackage{graphicx}
\usepackage{subcaption}
\usepackage[style=ieee,natbib=true,maxbibnames=8,url=false]{biblatex}
\usepackage{booktabs}
\usepackage{multirow}
\usepackage[bookmarks=true,hidelinks]{hyperref}
\usepackage{cleveref} 
\Crefformat{figure}{#2Fig.~#1#3}
\Crefmultiformat{figure}{Figs.~#2#1#3}{ and~#2#1#3}{, #2#1#3}{ and~#2#1#3}

\title{\LARGE \bf
Latent Action Diffusion for Cross-Embodiment Manipulation
}

\author{Erik Bauer$^{1,3}$ and Elvis Nava$^{1,2,3,4}$ and Robert K. Katzschmann$^{1,2,3,4}$
\thanks{$^{1}$mimic robotics, Zurich, Switzerland}%
\thanks{$^{2}$ETH AI Center, ETH Zurich, Zurich, Switzerland}%
\thanks{$^{3}$Soft Robotics Lab, Dept. of Mechanical and Process Engineering, ETH Zurich, Zurich, Switzerland}%
\thanks{$^{4}$Institute of Neuroinformatics, ETH Zurich and University of Zurich, Zurich, Switzerland}%
\thanks{\tt \footnotesize 
\{\href{mailto:erbauer@ethz.ch}{erbauer},
\href{mailto:enava@ethz.ch}{enava},
\href{mailto:rkk@ethz.ch}{rkk}\}@ethz.ch}%
\thanks{Project page: \url{https://mimicrobotics.github.io/lad/}}%
}

\let\oldtwocolumn\twocolumn
\renewcommand\twocolumn[1][]{%
    \oldtwocolumn[{#1}{
    \begin{center}
    \vspace{-20pt}
           \includegraphics[width=\textwidth]{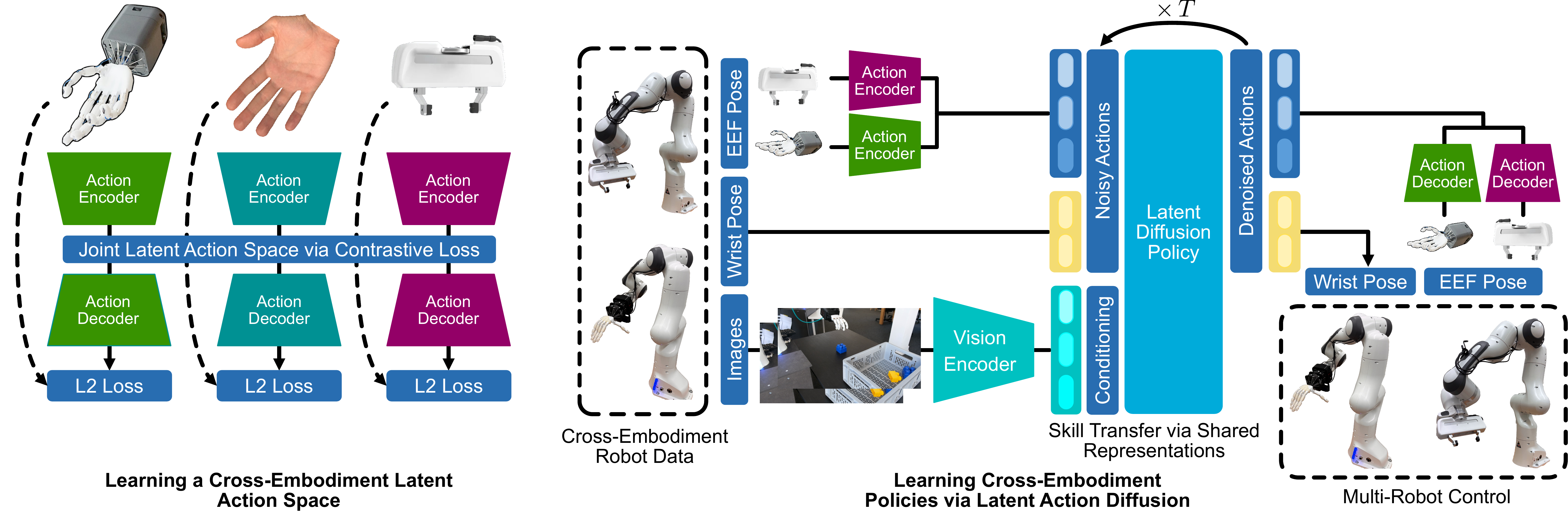}
           \captionof{figure}{Overview of our approach. \textbf{Left:} We construct a semantically aligned latent action space by training modality-specific encoders and decoders with a contrastive loss on retargeted end-effector (EEF) pose data from diverse end-effectors (dexterous hands, parallel gripper). \textbf{Right:} A single diffusion policy is trained in this shared latent space, enabling cross-embodiment policy learning, which in turn enables in multi-robot control with a single policy and skill transfer across embodiments.}
           \label{fig:title_figure}
        \end{center}
    }]
}

\begin{document}

\maketitle
\thispagestyle{empty}
\pagestyle{empty}

\begin{abstract}

End-to-end learning is emerging as a powerful paradigm for robotic manipulation, but its effectiveness is limited by data scarcity and the heterogeneity of action spaces across robot embodiments. In particular, diverse action spaces across different end-effectors create barriers for cross-embodiment learning and skill transfer. We address this challenge through diffusion policies learned in a latent action space that unifies diverse end-effector actions. We first show that we can learn a semantically aligned latent action space for anthropomorphic robotic hands, a human hand, and a parallel jaw gripper using encoders trained with a contrastive loss. Second, we show that by using our proposed latent action space for co-training on manipulation data from different end-effectors, we can utilize a single policy for multi-robot control and obtain up to 25.3\% improved manipulation success rates, indicating successful skill transfer despite a significant embodiment gap. Our approach using latent cross-embodiment policies presents a new method to unify different action spaces across embodiments, enabling efficient multi-robot control and data sharing across robot setups. This unified representation significantly reduces the need for extensive data collection for each new robot morphology, accelerates generalization across embodiments, and ultimately facilitates more scalable and efficient robotic learning.

\end{abstract}

\section{Introduction}
End-to-end learning is a promising path toward creating adaptable, generalist robots. Scaling up both the data volume and diversity to match the desired model capabilities is extremely resource-intensive and expensive, and inevitably requires pooling together data from different robotic embodiments. However, efficiently learning from different embodiments remains a significant challenge, as observation and action spaces vary significantly across robots (the ``embodiment gap``). 

Recent works on cross-embodiment learning have largely avoided explicitly addressing the problem of the embodiment gap in action spaces by only using data with a shared action space for pre-/co-training~\cite{Brohan2022RT1RT, Brohan2023RT2VM, octomodelteam2024octoopensourcegeneralistrobot}. Other works showing pretraining on human manipulation datasets have relied on explicitly aligning the human action space to the robot action space \cite{shaw2022videodexlearningdexterityinternet, yang2024pushinglimitscrossembodimentlearning, kareer2024egomimicscalingimitationlearning, wang2023mimicplaylonghorizonimitationlearning}. In this work, instead of using an explicit action space, we introduce a learned latent action space which can encode diverse action spaces from different end-effectors into a unified, semantically aligned latent action space. To achieve semantic alignment within the latent action space, we utilize retargeting methods, which enable precise alignment of different end-effector action spaces. For policy learning with latent actions, we factorize policies into an embodiment-agnostic policy trained on latent actions and multiple embodiment-specific decoders that are trained separately. Our proposed framework combines the simplicity of training policies with aligned observation and action spaces while still enabling learning from diverse robotic embodiments.

In particular, we focus on embodiment transfer among single-arm robots with different end-effectors. For our experiments, we utilize the Faive robotic hand~\cite{toshimitsu2023getting}, the mimic hand~\cite{nava2025mimiconescalablemodelrecipe} and a Franka parallel gripper. In two experiments, pairing data from each dexterous hand with data from the Franka gripper utilizing our proposed framework~(\Cref{fig:title_figure}), we compare latent diffusion policies co-trained on cross-embodiment data with single-embodiment diffusion policies. We demonstrate that our methodology enables both cross-embodiment control with a single policy and facilitates positive skill transfer, with up to 25.3\% performance (13.4\% average) improvement compared to single-embodiment diffusion policies, despite a significant embodiment gap. 

Our results indicate the potential of utilizing contrastive learning to bridge heterogeneous action spaces. As increasingly dexterous, human-like end-effectors become more common, our methodology provides a path forward for effectively sharing and reusing datasets across embodiments with diverse end-effectors through a unified latent action space.  

\begin{figure*}[ht]
    \centering
    \smallskip
    \smallskip
    \includegraphics[width=\linewidth]{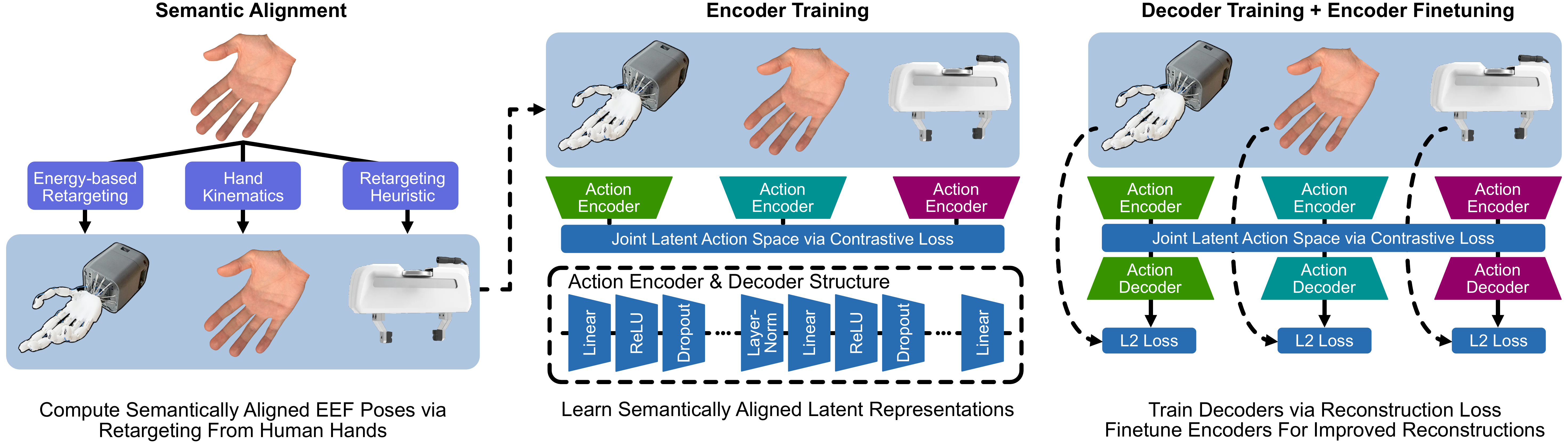}
    \caption{The three-stage process for learning the cross-embodiment latent action space. Stage 1: Aligned end-effector (EEF) poses are generated by retargeting human hand poses to different robot end-effectors. Stage 2: Embodiment-specific encoders are trained to project these actions into a shared latent space using a contrastive loss. Stage 3: Decoders are trained to reconstruct the original poses from the latent space, and encoders are fine-tuned to improve reconstruction quality.}
    \label{fig:four_stages}
\end{figure*}

\subsection{Contributions}

Our main contributions are:

\begin{enumerate}
    \item We introduce a general framework that unifies diverse end-effector action spaces into a single, semantically aligned latent space using contrastive learning, enabling downstream learning across diverse robotic embodiments.
    \item We show that factorizing diffusion policies into a latent, embodiment-agnostic policy and embodiment-specific action decoders enables multi-robot control across substantially different robot morphologies.
    \item We demonstrate substantial real-world performance gains — up to 25.3\% success rate improvement — from cross-embodiment co-training, and provide ablation studies validating our architecture for learning latent action spaces.
\end{enumerate}

\section{Related Works}

Learning from cross-embodiment data is a promising path towards scaling up both the volume and diversity of training data for robot learning that has been considered from various angles in prior works[cite: 74].

\paragraph{Constrained Action Space} \textit{RT-1}~\cite{Brohan2022RT1RT} showed positive skill transfer by co-training on multi-robot datasets with the same action space. Building on the Open-X-Embodiment collaboration~\cite{embodimentcollaboration2024openxembodimentroboticlearning}, multiple approaches have explored large-scale pretraining on more diverse robot data~\cite{Brohan2023RT2VM, octomodelteam2024octoopensourcegeneralistrobot, kim2024openvlaopensourcevisionlanguageactionmodel}. However, these approaches rely on constraining the action space for pretraining to a 7-dimensional space, effectively discarding any data with more complex action spaces, such as dexterous hands. In contrast, our approach utilizes a learned latent action space where the expressivity is a design choice detached from the physical constraints of any single embodiment.

\paragraph{One-Way Retargeting} 

\begin{figure*}
    \centering
    \smallskip
    \includegraphics[width=0.9\linewidth]{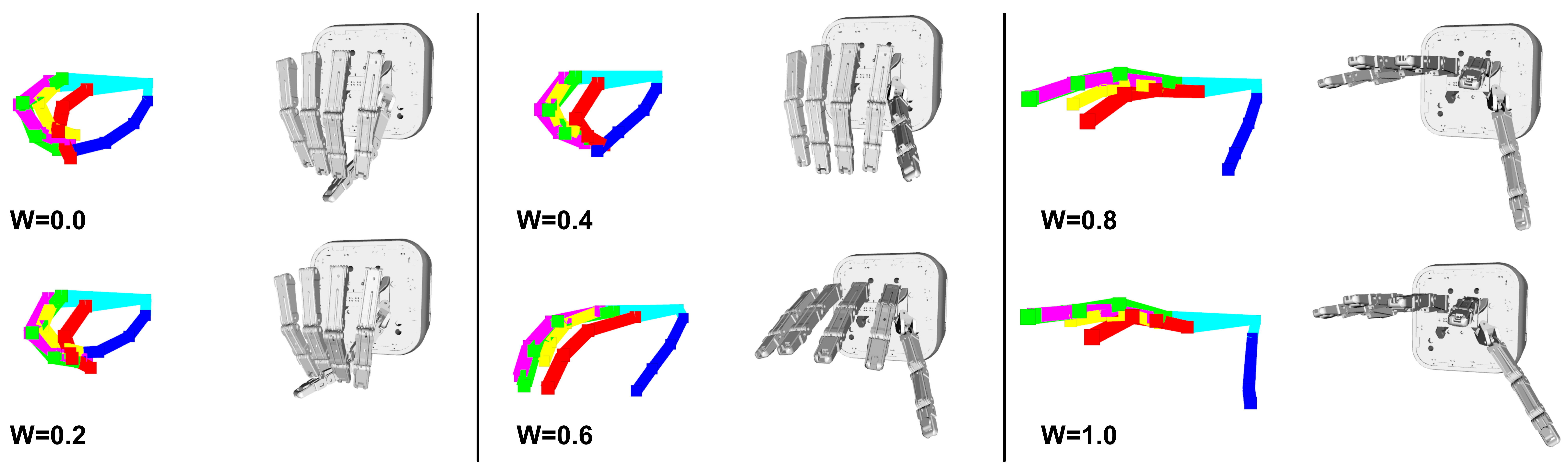}
    \caption{Qualitative evaluation of the joint latent action space. We encode normalized gripper widths \(W \in [0,1]\) (from closed to open) and perform cross-modal reconstruction by decoding them into human hand poses (colored lines on left) and poses for the Faive hand (grey model on right). Existing approaches using retargeting only allow for single-directional retargeting (i.e. human hands to robot hands), which is a limitation our latent action space overcomes. Any modality can be encoded and decoded to any other modality under the alignment constraints of the data.} 
    \label{fig:contrastive_retargeting}
\end{figure*}

Multiple approaches that focus on skill transfer from human video~\cite{shaw2022videodexlearningdexterityinternet, wang2023mimicplaylonghorizonimitationlearning, kareer2024egomimicscalingimitationlearning} propose retargeting human actions to the action space used by their robot. These approaches limit the utility of the resulting policy to the specific embodiment for which the retargeting was designed, as the policy predicts actions specific to that embodiment. Our methodology overcomes this bottleneck by learning a shared, semantically aligned latent action space that uses retargeting as a prior for alignment. This unified action representation supports any-to-any reconstruction across heterogeneous end-effectors~(\Cref{fig:contrastive_retargeting}), enabling policies trained with latent actions to control multiple embodiments.

\paragraph{Alignment-Free Methods} 
Different approaches have investigated cross-embodiment learning without explicit alignment between different action spaces~\cite{black2024pi0visionlanguageactionflowmodel, doshi2024scalingcrossembodiedlearningpolicy}. However, skill transfer is largely driven by data scale for these architectures, as they lack architectural inductive biases to encourage transfer. The underlying assumption of large data volumes is often not satisfied for dexterous end-effectors with high-dimensional action spaces. Our proposed latent action space uses retargeting as an alignment prior that facilitates skill transfer between complex action spaces without requiring large data volumes. 

Other methods propose video representations as embodiment-agnostic actions~\cite{ye2024latentactionpretrainingvideos, chen2024moto}, but these capture coarse motion primitives and require embodiment-specific finetuning before deployment. By learning a latent space directly from explicit action spaces instead of video, our approach encodes precise low-level actions, enabling deployment of the same model on different robots without further finetuning.

\section{Methodology}

\noindent Our proposed framework for cross-embodied policy learning consists of two parts: learning a latent action space~(\Cref{fig:four_stages}) and training latent policies. In this work, we focus on policy learning for single-arm robots with different end-effectors in a unified latent action space. 

Our key insight is that learning aligned representations for different end-effector action spaces can be viewed as a multimodal representation learning problem. Based on this perspective, we design a pipeline for cross-embodied latent imitation learning comprised of the following steps: generating paired action data, learning encoders and decoders for the shared latent space, and latent policy learning.

\subsection{Creating Aligned Action Pairs}
\noindent Multimodal representation learning architectures for \(M\) modalities generally rely on tuples containing paired data of the form \(\mathbf{x}_i = \left( x_i^1, x_i^2, \dots, x_i^M \right) \), where there is some form of cross-modal correspondence between the elements of each tuple. In multimodal learning, correspondences between data modalities are typically created through manual annotation (\textit{e.g.}, image-caption pairs~\cite{sutter2021generalizedmultimodalelbo}) or created with modality-specific expert models (\textit{e.g.}, creating depth pseudolabels from RGB images~\cite{mizrahi20234mmassivelymultimodalmasked}). 

In the context of robotics, we are looking for alignment functions between different action spaces which allow us to establish mappings in between the action spaces. We focus on aligning action spaces of different end-effectors (human hands, anthropomorphic robotic hands, parallel jaw gripper, \dots). For this subproblem, retargeting functions from human hands to robotic end-effectors are a useful prior for alignment, as they typically already exist for different embodiments in order to teleoperate robots.

To construct tuples of paired end-effector poses, we proceed as follows:

\begin{equation}
    \mathbf{x}_i = (x_i^H, f_{H}^{R_1} \left( x_i^{H} \right), \dots, f_{H}^{R_M} \left( x_i^{H} \right))
\end{equation}

\noindent where \( f_{H}^{R_j}, \ j \in \{1, \dots, M\} \) are retargeting functions from human hands to the j-th robot embodiment.

\subsubsection{Action Representations} 

For human hands, we derive a 189-dimensional pose representation \(\mathbf{\theta}_H\) using the local transformations in between the 21 joints according to the kinematic chain of the hand. To represent rotations, we utilize the continuous 6D rotation representation proposed by \cite{zhou2020continuityrotationrepresentationsneural}. Poses for the Faive hand or mimic hand are represented as an 11- or 16-dimensional vector of joint angles \( \mathbf{\theta}_F\) or \( \mathbf{\theta}_M\) respectively. Poses of parallel jaw grippers are represented as normalized one-dimensional gripper width \( \theta_P \in [0,1] \).

\subsubsection{Retargeting} \label{sec:retargeting}

For retargeting, we follow the technique introduced by \cite{sivakumar2022robotictelekinesislearningrobotic}, which utilizes keyvectors for both the human and robot hand. The keyvectors \(v_i^{\{H,M\}} (\mathbf{\theta}_{\{H,M\}})\) are vectors from the palm to each fingertip and from each fingertip to all other fingertips and provide a unifying representation that can be defined for any hand with a notion of fingertips. To map from human hands to complex robotic hands such as the mimic hand, we can formulate retargeting functions as a minimum-energy solution to the squared keyvector difference of the human and the robot hand.  By using the forward kinematics of each hand, we can determine the keyvectors as a function of its respective pose representation \(\mathbf{\theta}_H\) or \(\mathbf{\theta}_H\). As a concrete example, to retarget from a human hand pose \(\mathbf{\theta}_H\) to a mimic hand pose \(\mathbf{\theta}_M\), we can directly optimize over the \(\mathbf{\theta}_M\) with the differentiable objective shown in~\Cref{eq:retargeting_mimic}. Each pair of keyvectors has a scaling factor \(s_i\), which is used to compensate for different finger lengths. For all 15 keyvectors, scaling factors are determined through qualitative evaluation. The resulting retargeting function can be expressed as follows:

\begin{equation}
\label{eq:retargeting_mimic}
    \mathbf{\theta}_M (\mathbf{\theta}_H) = \text{argmin}_{\mathbf{\theta}_F} \sum_{i=1}^{15} \left| \left| v_i^H(\mathbf{\theta}_H) - s_i v_i^F(\mathbf{\theta}_M) \right| \right|_2^2
\end{equation}

\noindent For parallel jaw grippers, we take the minimum of all keyvectors originating at the thumb and normalize it by a standard gripper width \(W\) such that \(\theta_P \in [0,1]\):

\begin{equation}
    \theta_P (\mathbf{\theta}_H) = \min_{\theta_P} \left( \min_i \frac{\left| \left| v_i^H (\mathbf{\theta}_H) \right| \right| }{W}, 1 \right)
\end{equation}

To add other robotic end-effectors to the learning scheme, it is only necessary to find a retargeting function from either human hands or another robotic end-effector to the newly added one.

\subsection{Contrastive Latent Space Learning}

\noindent For a shared latent action space, it is crucial that 1) for each modality, sufficient information is encoded such that we can precisely reconstruct end-effector poses and 2) the latent space has a coherent structure, meaning that the cross-modal alignment present in the model inputs during training is upheld in the learned latent space. To achieve both of these goals, we propose a two-step learning procedure: first, using batches with \(B\) aligned end-effector poses that were generated via retargeting, \(M\) modality-specific encoders \(q_m, m \in 1 \dots M\) are trained that project actions \(x_m\) from each input modality into a shared latent space, where we utilize a pairwise InfoNCE loss~\cite{oord2019representationlearningcontrastivepredictive} to ensure alignment within the batch:

\begin{align}
    \begin{split}
    &\mathcal{L}_\text{contrastive} = \frac{1}{M(M-1)} \\&\sum_{i=1}^M \sum_{j=i+1}^M \left(-\frac{1}{B} \sum_{n=1}^B \log \frac{\exp(q_i(x_i^n) \cdot q_j(x_j^n)/\tau)}{\sum_{k=1}^B \exp(q_i(x_i^n) \cdot q_j(x_j^k)/\tau)}\right)
    \end{split}
\end{align}

\noindent where \(\tau\) denotes the temperature. In the second stage, we train \(M\) modality-specific decoders \(p_m, m \in 1 \dots M \), which learn to reconstruct ground truth actions \(\Hat{x}_i\) from their latent representations. Additionally, the encoders \(q_m\) are fine-tuned with a lower learning rate. The total loss \(\mathcal{L}_{\text{total}}\) backpropagated through the encoders and decoders is a combination of a reconstruction loss \(\mathcal{L}_{\text{recon}}\) and the previous contrastive loss \(\mathcal{L}_\text{contrastive}\), where the hyperparameter \(\lambda\) can be used to control the trade-off in between alignment and self-reconstruction.

\begin{align}
    \mathcal{L}_{\text{recon}} &= \frac{1}{M} \sum_{i=1}^M \sum_{n=1}^B \left| \left| p_i\left(q_i\left(x_i^n\right)\right) - \Hat{x}_i^n \right| \right|_2^2 \\
    \mathcal{L}_{\text{total}} &= \mathcal{L}_{\text{recon}} + \lambda \mathcal{L}_\text{contrastive}
\end{align}

\noindent We use a cross-reconstruction (CR) loss for validation. From modality \(i\) to \(j\), given paired end-effector poses \((x_i^n, x_j^n)\), the CR-Loss is: 
\begin{equation}
    \mathcal{L}_{\text{CR(i,j)}} = \frac{1}{B} \sum_{n=1}^B \left| \left| p_j\left(q_i\left(x_i^n\right)\right) - x_j^n \right| \right|_2^2
\end{equation} 

\noindent We encode data from modality \(x_i^n\), decode it to modality \(j\) and evaluate the result versus the paired ground truth data \(x_j^n\). The self-reconstruction (SR) loss used for validation is \(\mathcal{L}_{\text{SR(i)}}=\mathcal{L}_{\text{CR(i,i)}}\).

Action decoders and encoders are parameterized by standard multi-layer-perceptrons (MLPs) as encoders and decoders~(\Cref{fig:four_stages}). The input layer consists of a linear layer, ReLU layer and a dropout layer. For each hidden layer, we first have a normalization layer, then a linear layer, followed by a ReLU (rectified linear unit) activation and a dropout layer. After the hidden layers, a last linear layer is used to project the output to the desired dimension.

\subsection{Policy Learning}

\noindent With a learned latent action space, we employ Diffusion Policy~\cite{chi2024diffusionpolicy} to map shared observations across datasets to latent end-effector actions of different embodiments and non-latent wrist poses of the robot arm as shown in~\Cref{fig:title_figure}. The encoders and decoders of the contrastive action model stay frozen for latent policy learning, with the diffusion objective being applied on the denoised latent end-effector actions and wrist poses. For inference, the suitable embodiment-specific decoder is used to obtain a decoded end-effector pose from the output of the latent policy. 

We validate our method across two diffusion policy implementations: a transformer-based implementation~\cite{robomimic2021} and a U-Net-based implementation~\cite{chi2024diffusionpolicy}. The transformer-based policy is utilized in experiments with the Faive hand and Franka gripper, whereas the U-Net-based implementation is used for experiments with the mimic hand and the Franka gripper. Specifications for the models used in the experiments are given in \Cref{tab:diffusion_policy_hyperparameters}.

For co-training on differently sized datasets, we assign normalized weights \(w_j\) to all datasets. During training, we seek to combine samples from all datasets to fill batches with \(B\) samples in total. We sample per-dataset sub-batches with appropriately rounded sizes \(\text{round}(\frac{B}{w_j})\), project the actions into the shared latent action space, normalize the sub-batches, and then concatenate them into a single batch for efficient training. Through this mechanism, the weight of each dataset approximately represents a sampling probability for each training step. In our experiments, all datasets have the same weights, such that the model is exposed to equal amounts of data from all embodiments.

\begin{table}[ht]
	\centering 
	\caption{Diffusion Policy Specifications} \label{tab:diffusion_policy_hyperparameters} 
	\begin{tabular}{l l l} \toprule \textbf{Hyperparameter} & \textbf{Transformer} & \textbf{U-Net} \\
    \midrule
    Parameter Count & \(\sim\)40M & \(\sim\)44M \\
    Vision Encoder & Resnet18 & ViT-S\\
    Batch Size & 300 & 256 \\
    Horizon & 21 timesteps & 48 timesteps \\
    Diffusion Noise Schedule & Squared cosine & Squared cosine \\ $\beta_{\text{start}}$ & 0.0001 & 0.0001 \\$\beta_{\text{end}}$ & 0.02 & 0.02 \\
    Peak Learning Rate & 0.0001 & 0.0001 \\
    Optimizer & AdamW & AdamW \\
    Training Steps & 90k & 40k \\
    \bottomrule
    \end{tabular}
\end{table}

\begin{table}[ht]
\centering
\caption{Contrastive Action Model Specifications}
\label{tab:contrastive_hyperparameters}
\begin{tabular}{l l}
\toprule
\textbf{Hyperparameter} & \textbf{Value} \\
\midrule
Parameter Count (single encoder/decoder) & \(\sim\)26k \\
Batch Size & 16384 \\
Learning Rate & 0.00001 \\
Finetuning Learning Rate & 0.0001 \\
Optimizer & AdamW \\
Weight Decay & 0.001 \\
Temperature Schedule & Exponential\\
Temperature Start\(\rightarrow\)End & 0.25\(\rightarrow\)0.16 \\
Latent Space Dimension & 16 \\
\(\lambda\) & 0.1 \\ 
MLP Hidden Dimensions & 32, 128, 128, 32 \\
Training Epochs (Encoders) & 5000 \\
Training Epochs (Decoders + finetune encoders) & 10000 \\
\bottomrule
\end{tabular}
\end{table}

\begin{table}[t]
\smallskip
\centering
\caption{Ablation study on the impact of temperature annealing (TA) and finetuning (FT) on the contrastive action model, evaluating self- and cross-reconstruction losses.}
\label{tab:ablation_study}
\begin{tabular}{lccccc}
\toprule
\multirow{2}{*}{\textbf{Method}} & \multicolumn{2}{c}{\textbf{SR-Loss}} & \multicolumn{2}{c}{\textbf{CR-Loss}} \\
\cmidrule(lr){2-3} \cmidrule(lr){4-5}
 & mimic & Franka & mimic\(\rightarrow\)Franka & Franka\(\rightarrow\)mimic\\
\midrule
Full (ours) & \textbf{0.762} & 3.7e-8 & \textbf{0.002} & \textbf{214.20} \\
no TA & 0.948 & \textbf{1.5e-8} & 0.007 & 286.64 \\
no FT & 44.76 & 2.6e-8 & 0.013 & 391.85 \\
no FT\&TA & 49.765 & 2.1e-8 & 0.02 & 397.23 \\
\bottomrule
\end{tabular}
\end{table}

\begin{figure*}
    \centering
    \includegraphics[width=\linewidth]{figures/success_rates_wide_new.pdf}
    \caption{Success rates for three different tasks comparing single-embodiment diffusion policies to cross-embodied latent diffusion policies trained on data from both embodiments for each task. Block stacking: 200 demos per embodiment, one external camera. Block pick and place: 200 demos per embodiment, one external camera + wrist camera for mimic hand, replaced by zero-padding for Franka gripper. Plush toy pick and place: 100 demos per embodiment, one external camera.}
    \label{fig:success_rates}
\end{figure*}

\begin{figure*}
    \centering
    \includegraphics[width=\linewidth]{figures/rollouts_wide_captions.pdf}
    \caption{Cross-embodiment policy rollouts for three tasks (block stacking, block pick and place, plush toy pick and place). For each task, both embodiments are controlled by the same cross-embodiment diffusion policy, demonstrating multi-robot control.}
    \label{fig:rollouts}
\end{figure*}

\section{Experimental Results and Discussion}

We conducted experiments covering three different end-effectors and three tasks across two setups: one with the Faive hand and the Franka gripper and one setup with the mimic hand and the Franka gripper. For each end-effector in each setup, we compare single-embodiment policies with one cross-embodiment policy co-trained on data from all end-effectors. We train policies using contrastive action models that were trained for the end-effectors present in each task. In addition, we validate our design choices for our contrastive action model through an ablation study.

\subsection{Ablation Study: Contrastive Action Model}
To validate our design choices, we compare several versions of the contrastive action model~(\Cref{tab:ablation_study}). As metrics, we utilize self-reconstruction (SR) and cross-reconstruction (CR) validation losses. The losses shown in the table are denormalized losses for the 16-dimensional action space of the mimic hand and the 1-dimensional action space of the Franka gripper. Due to the higher dimension and complexity of the larger action space, the losses for the mimic hand are larger than the losses for the Franka gripper.

We compare the full training pipeline for the contrastive model to three ablations. The ablation without temperature annealing keeps the temperature constant at the previous final value. The ablation without finetuning freezes the encoders in the second training step while the decoders are being trained. Both temperature annealing and finetuning the encoders reveal themselves to substantially improve both self- and cross-reconstruction metrics, with finetuning being the most important addition to the pipeline.

\subsection{Experimental Setups}

We evaluate our framework across three manipulation tasks using different combinations of end-effectors. For the Block Stacking task, we use the mimic hand and Franka gripper to pick a block, stack it on another, and place the pair into a box. For this task, we collected 200 demonstrations per embodiment using a single external camera for visual observations. Each policy is evaluated over 70 trials.

For the Block Pick \& Place task, we use the mimic hand and the Franka gripper to place a plastic cube inside a box. We collected 200 demonstrations for each embodiment. In addition to an external camera and the robot arm pose, this setup uses a wrist camera for the mimic hand, which is replaced by zero-padding for the Franka gripper to test learning with asymmetric observations. Each policy is evaluated over 25 trials.

Finally, for the Plush Toy Pick \& Place task, we use a Franka parallel gripper and a dexterous Faive hand to pick up a plush toy and place it into a bowl. For this setup, we collected 100 demonstrations for each end-effector and used a single external RGB camera for observations. Each policy is evaluated over 40 trials.

\subsection{Results and Discussion}

Our experiments demonstrate that co-training policies in a learned latent action space enable both multi-robot control from a single policy and significant performance gains through cross-embodiment skill transfer. We analyze the performance of our co-trained policies against single-embodiment baselines across three distinct manipulation tasks.

The multi-stage Block Stacking task showed the most pronounced skill transfer. For the initial coarse manipulation stage ("Pick block"), the co-trained policy yielded absolute success rate improvements of 25.3\% for the mimic hand and 13\% for the Franka gripper. More notably, the Franka gripper performance in the subsequent fine-grained stages ("Stack blocks" and "Put into box") improved significantly by 13\% and 11\%, respectively. This indicates that the gripper effectively learned more precise manipulation strategies from the dexterous hand data. The mimic hand saw a slight performance decrease in the final stage, likely due to the hand occluding the block from the camera after grasping it, making the final placement more difficult.

For the Block Pick \& Place task, which featured asymmetric camera observations, the benefits of co-training were still evident for the more sensor-rich embodiment. The success rate of the mimic hand improved by 13\% over its single-embodiment baseline. However, the Franka gripper's performance decreased. We attribute this to the policy becoming reliant on the wrist camera view available only to the mimic hand, highlighting that skill transfer in the presence of asymmetric observations remains a significant challenge, despite the aligned action space.

In the Plush Toy Pick \& Place task, the co-trained policy substantially outperformed the single-embodiment versions, achieving a 10\% higher success rate for the Faive hand and a 7.5\% improvement for the Franka gripper. This suggests that the policy learns a more robust, shared representation of the task, allowing skills learned from the dexterous hand (e.g., object handling) to benefit the simpler gripper, and vice versa.

In summary, our approach successfully unifies control across diverse robotic hardware and facilitates cross-embodiment skill transfer. The results confirm that policies learn helpful shared representations, leading to improved success rates in both coarse and fine-grained manipulation.

\subsection{Limitations}

\paragraph{Asymmetric Observations} Our policy struggles to achieve consistent performance on all embodiments if they have different sensor modalities (e.g., wrist camera for one robot only). This remains a fundamental challenge for cross-embodiment learning. Future work could explore vision encoder adaptation or learned observation alignment.

\paragraph{Latent Space Regularization} Our latent space is not explicitly regularized, which may lead to non-smooth regions that are harder to model for downstream policies. Integrating VAE-style priors or enforcing smoothness via time-contrastive losses could guarantee desirable latent space properties.

\paragraph{Dataset  Scale Imbalance} We observe limited gains when adding large-scale datasets (e.g., BridgeV2~\cite{walke2023bridgedata}, DexYCB~\cite{chao:cvpr2021}) to our co-training mix. Training on datasets with highly imbalanced sizes and transferring skills despite significant environmental differences remains a fundamental challenge of cross-embodiment learning that could be explored via more sophisticated sampling strategies to determine dataset weights and visual representation learning that facilitates skill transfer given largely disjunct observation spaces.

\section{Conclusion}
\label{sec:conclusion}

\noindent Among current challenges in robotics, enabling effective skill transfer across diverse embodiments is of crucial importance to both maximize the volume and diversity of suitable training data and to ensure the reusability of training data throughout the life cycle of different end-effectors. To this end, we frame cross-embodiment learning with different end-effectors as a multimodal representation learning problem and propose a two-stage pipeline to learn capable policies that can control multiple end-effectors. In real-world experiments with dexterous hands and a Franka parallel gripper, we demonstrate that through co-training on cross-embodiment data with our method for latent action spaces, we enable both multi-robot control and positive skill transfer across embodiments. In particular, the performance improvement of up to 25.3\% (average: 13.4\%) indicates that our method facilitates skill transfer between end-effectors with a large embodiment gap and underlines its potential for wider use across a broader range of robot morphologies. Future work includes expanding our method to a more diverse ecosystem of end-effectors and further investigating the behavior of skill transfer across different dataset sizes with more distinct visual differences.





\section*{ACKNOWLEDGMENT}

We express our gratitude to Emanuele Palumbo for valuable discussions and insights on multimodal learning architectures. Furthermore, we are grateful for support from Chenyu Yang in setting up the initial version of \href{https://github.com/srl-ethz/srl_il}{srl\_il} and Davide Liconti in writing the control interface for the Franka gripper. Additionally, we thank the mimic team for support in realizing experiments with the mimic hand: Victor Montesinos, Jonas Pai, Benedek Forrai, Robert Malate, Philipp Wand, Stephan Polinski, and Norica Bacuieti.

We are grateful for grant funding from Innosuisse (122.679 SIP) and ESA BIC received by mimic robotics in connection to this project. We are also grateful to funding supporting academic co-authors: funding from the ETH AI Center, SNSF Project Grant MINT \#200021\_215489, Swiss Data Science Center, Amazon Research Award, Armasuisse.


\addtolength{\textheight}{-5cm}   
\printbibliography[]

\end{document}